\documentclass[letterpaper]{article} 
\usepackage{aaai25}  
\usepackage{times}  
\usepackage{helvet}  
\usepackage{courier}  
\usepackage[hyphens]{url}  
\usepackage{graphicx} 
\usepackage{natbib}  
\usepackage{caption} 
\usepackage{algorithmicx, algorithm}
\usepackage[noend]{algpseudocode}
\usepackage{appendix}
\usepackage{amsmath}
\usepackage{amssymb}
\usepackage{booktabs}

\usepackage{newfloat}
\usepackage{listings}
\DeclareCaptionStyle{ruled}{labelfont=normalfont,labelsep=colon,strut=off} 
\floatstyle{ruled}
\newfloat{listing}{tb}{lst}{}
\floatname{listing}{Listing}
\title{Energy-Guided Optimization for Personalized Image Editing with Pretrained Text-to-Image Diffusion Models}
\author{
    Rui Jiang\textsuperscript{\rm 1}\equalcontrib, Xinghe Fu\textsuperscript{\rm 1}\equalcontrib, Guangcong Zheng\textsuperscript{\rm 1}, Teng Li\textsuperscript{\rm 1},
    Taiping Yao\textsuperscript{\rm 2},  Xi Li\textsuperscript{\rm 1}\thanks{Corresponding author.}
}
\affiliations{
    \textsuperscript{\rm 1}College of Computer Science and Technology, Zhejiang University\\
    \textsuperscript{\rm 2}Youtu Lab, Tencent\\

    \{jrss, xinghefu, guangcongzheng, tengli19\}@zju.edu.cn, taipingyao@tencent.com, xilizju@zju.edu.cn
}

\usepackage{bibentry}

\begin{document}

\maketitle

\begin{abstract}
The rapid advancement of pretrained text-driven diffusion models has significantly enriched applications in image generation and editing. However, as the demand for personalized content editing increases, new challenges emerge especially when dealing with arbitrary objects and complex scenes. Existing methods usually mistakes mask as the object shape prior, which struggle to achieve a seamless integration result. The mostly used inversion noise initialization also hinders the identity consistency towards the target object. To address these challenges, we propose a novel training-free framework that formulates personalized content editing as the optimization of edited images in the latent space, using diffusion models as the energy function guidance conditioned by reference text-image pairs. A coarse-to-fine strategy is proposed that employs text energy guidance at the early stage to achieve a natural transition toward the target class and uses point-to-point feature-level image energy guidance to perform fine-grained appearance alignment with the target object.
Additionally, we introduce the latent space content composition to enhance overall identity consistency with the target. Extensive experiments demonstrate that our method excels in object replacement even with a large domain gap, highlighting its potential for high-quality, personalized image editing.
\end{abstract}

%

\section{Introduction}

With the rapid advancement of pretrained text-driven diffusion models \cite{Rombach2022, Ramesh2022, Saharia2022}, the field of image generation has experienced unprecedented growth in guided image manipulation techniques through textual directives.
The development of pretrained text-to-image (T2I) generation models has significantly enriched various domains, particularly image editing \cite{Mou2024, Mou2023, Mokady2023}, offering novel approaches to understanding and manipulating images.
However, as the demand for personalized content editing increases, new challenges arise, particularly when dealing with arbitrary objects and complex scenes.
Existing methods primarily focus on personalized image generation \cite{Gal2022, Ruiz2023}, aiming to create new images with customized content.

\begin{figure}[t]
    \centering
    \includegraphics[width=\columnwidth]{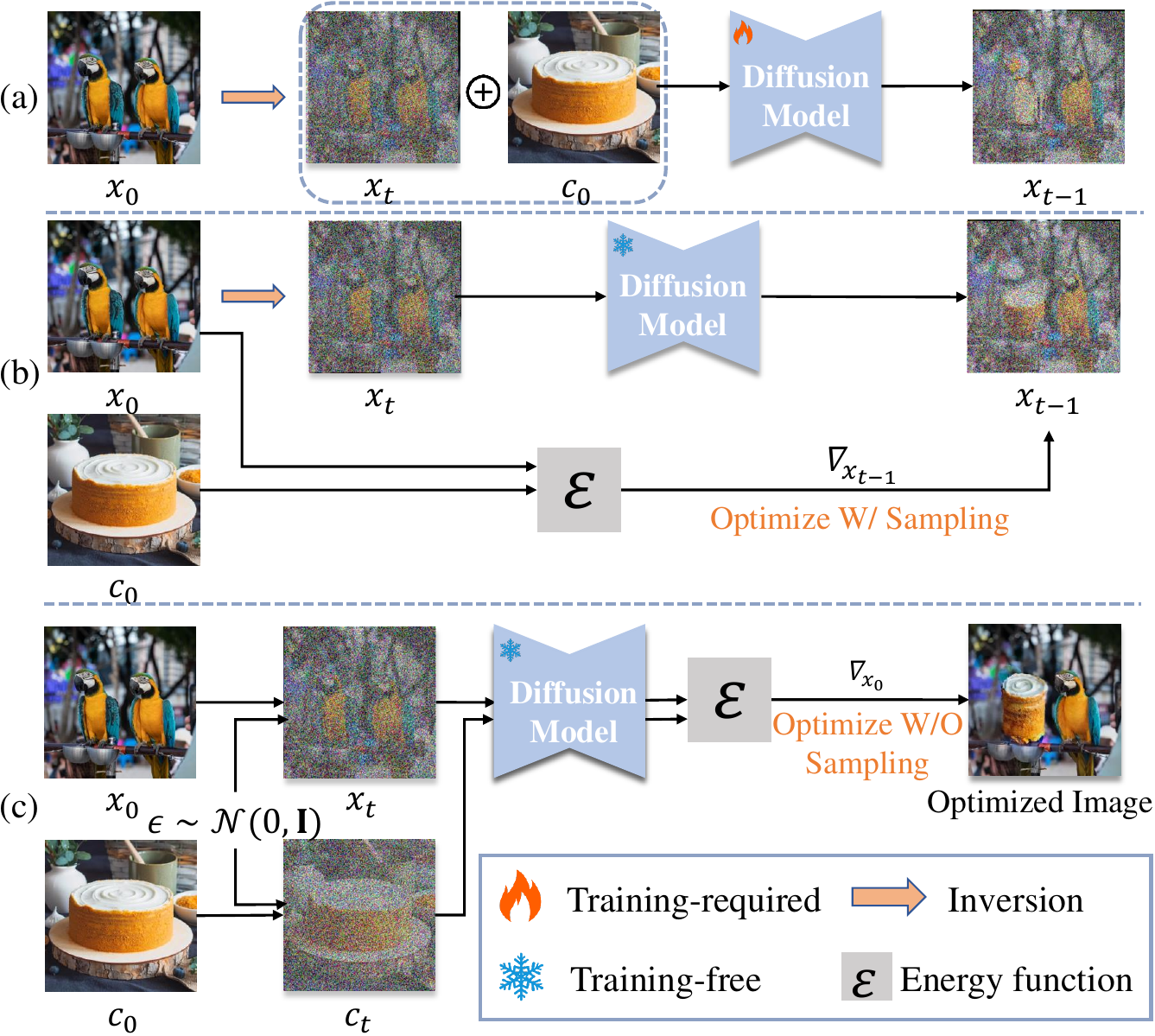}
    \caption{
    Comparisons among different methods in personalized content editing.
    (a) Inpainting-based methods usually require fine-tuning the diffusion model with reference images as the condition. (b) Sampling-based methods initialize the noise with inversion to maintain the background information from the source image. (c) The proposed method iteratively optimizes the latent code to perform training-free and inversion-free editing.}
    \label{fig:first}
\end{figure}

Personalized image editing involves placing a target object into a desired position within a scene image or replacing objects in a source image with those from a reference image. The primary challenges in this task are accurately integrating personalized content harmoniously into the target image and maintaining editing flexibility.

\begin{figure*}[t]
    \centering
    \includegraphics[width=0.85\linewidth]{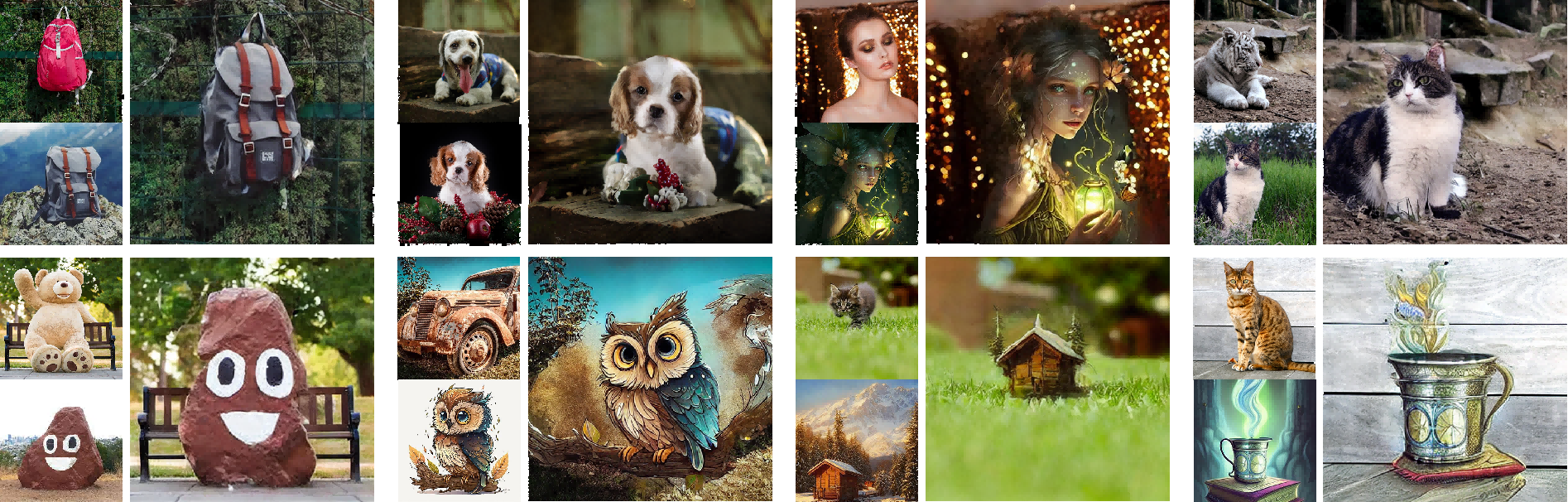}
        \caption{
        Performance overview of the proposed method in image customization editing.
        Our method generates edited images by integrating contextual guidance with a reference image. The first row demonstrates object replacement within the same category, while the second row shows object replacement across different categories.}
        \label{fig:teaser}
\end{figure*}

To address these challenges, there are two main categories of personalized image editing methods: inpainting-based and sampling-based (as shown in Fig.~\ref{fig:first}). These methods attempt to remove and regenerate objects to edit specific image regions. Other methods like SelfGuidance \cite{Epstein2023} and Diffeditor \cite{Mou2024} provide guidance in the sampling process and manipulate the sampling direction based on reference images. 

Previous methods achieve impressive performance in some applications (\textit{e.g.}, appearance replacement). However, some hard cases in personalized editing (\text{e.g.}, cross-class object replacement) remain challenging for previous methods. Most inpainting-based methods like Anydoor~\cite{Chen2024a} require extra training and treat the source mask as the target object shape prior. These cause unstable editing results when the shape mismatches between source and target objects or the test domain shifts. Sampling-based methods rely on the inversion of latent codes to maintain the background information from the source image. The sampling process (ODE or SDE) encounters the tradeoff between maintenance and flexibility, and makes it hard to present the target object in the edited image. Therefore, developing a training-free and inversion-free algorithm for personalized editing is in demand.

Unlike previous methods, we propose optimizing the latent code directly to obtain the edited image.
Score Distillation Sampling (SDS) \cite{Poole2022} and Delta Denoising Score (DDS) \cite{Hertz2023} utilize the pre-trained T2I diffusion model to control the optimization of the latent code and achieve text-conditioned editing. Inspired by this, the reference image in personalized editing can also be treated as a condition in optimization. The optimization is training-free and allows more flexibility in the edited area even for hard cases.

Our approach formulates personalized content editing as an energy-based optimization problem conditioned by a reference text-image pair.
First, we use the reference text and image as queries and the diffusion model as a conditioned energy function. The corresponding feature-level differences between the edited and reference images within the target object are minimized during optimization. 
Second, to achieve higher identity consistency for the target object, we design a replacement-based content composition operation that integrates the target information into the latent variables during optimization. Third, to avoid blurred output and achieve coarse-to-fine optimization, we propose scheduling the timesteps for the diffusion model in descending order and using text guidance only in early iterations. This allows a natural transition of the object shape towards the target and achieves stable refining of appearance at the late stage. Additionally, we utilize truncation and smoothing techniques for the gradients to maintain the background information and ensure a harmonious integration of the target object.

In summary, the contributions of this paper are as follows:
\begin{itemize}
\item[$\bullet$] We first conceptualize the problem of personalized image editing as a conditioned optimization task using diffusion-based text-image energy guidance. 
\item[$\bullet$] We propose an energy-guided optimization framework (EGO-Edit) involving coarse-to-fine strategies along with latent space content composition for stability and higher consistency in optimization.
\item[$\bullet$] Extensive experiments demonstrate the effectiveness of our method in personalized image editing, such as object swapping and inpainting, and can produce desired edited results in hard cases.
\end{itemize}

\begin{figure*}[!t]
  \centering
  \includegraphics[width=\textwidth]{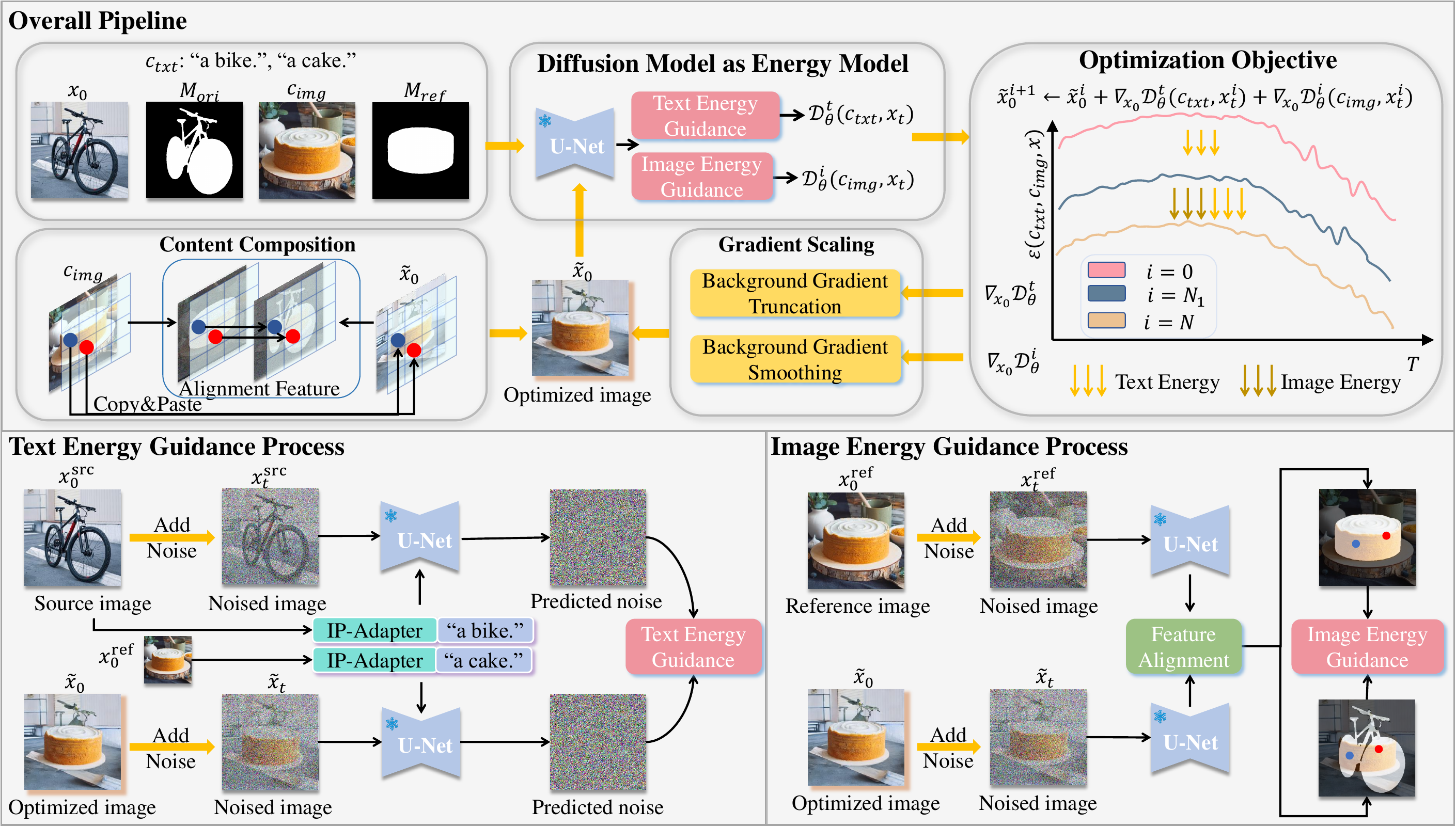}
  \caption{Pipeline Overview of the Proposed Method: The illustration above outlines the pipeline for our energy-guided optimization method. We construct the energy function derived from the diffusion model, aiming to minimize the energy of the edited image, \(\mathbf{\tilde{x}}_t\), to progressively align its distribution with that of the reference image. The diffusion-based energy function is composed of two key components: Text Energy Guidance (TEG) and Image Energy Guidance (IEG). TEG is applied throughout the entire process, ensuring consistent semantic alignment, while IEG is specifically employed during the N2 optimization step to refine visual details, enhancing the fidelity of the edited image to the reference. The processes for both TEG and IEG are detailed below the main pipeline.
}
  \label{pipeline}
\end{figure*}

\section{Related Work}
\label{Related Work}

\paragraph{Text-to-image Model.}
The field of text-to-image synthesis has been profoundly influenced by conditional diffusion models \cite{Dhariwal2021, Ho2022, Nichol2021}, which have ushered in a new era of image generation. These models have demonstrated the ability to produce high-quality images that are conditioned on textual descriptions, significantly advancing the capabilities of generative systems. However, the sensitivity of these models to the quality of the text prompt has become a recognized limitation, often necessitating careful and deliberate prompt design \cite{Hao2024} to achieve satisfactory results \cite{Witteveen2022}.
Recent works have begun to explore the integration of image prompts to guide the generation process. DALL-E 2 \cite{Ramesh2022} marked a pivotal step with its pioneering approach to image-guided image generation. Subsequent research, such as ELITE \cite{Wei2023}, BilpDiffusion \cite{Li2024}, ProFusion\cite{Zhou2023}, Domain-Agnostic\cite{Arar2023} and IP-Adapter \cite{Ye2023}, has focused on learning from image prompts to enable object customization, offering a new dimension in the control over generative models \cite{Jiang2024, wu2024customcrafter,realcam}.
Despite the recent strides in conditional image generation, leveraging the combined power of text and image prompts to enhance the versatility and precision of diffusion models in image editing tasks continues to be an open challenge.

\paragraph{Image Editing.} 
The image editing methodologies can be categorized into three directions:
training-based approaches \cite{Kim2022, Wang2023, Yang2024, Li2023, Sheynin2023}, test-time fine-tuning approaches \cite{Choi2023, Mokady2023, Dong2023}, and traing-free approaches \cite{Hertz2022, Parmar2023, Tumanyan2023, Lu2023}. 
Most of the previous editing \cite{Ju2024, HubermanSpiegelglas2024, Brooks2023} focus on editing local image regions conditioned by text prompts.

Paint-by-Example (PbE) \cite{Yang2023a} proposed a training-based approach for exemplar-guided personalized image editing.
Subsequent training-based methods Custom-Edit \cite{Choi2023}, ObjectStitch \cite{Song2023}, Uni-paint \cite{Yang2023}, DreamInpainter \cite{Xie2023}, Photoswap \cite{Gu2024} and Anydoor \cite{Chen2024a} also implemented personalized object replacement by inpainting.
Blip-diffusion \cite{Li2024} used a pre-trained multimodal encoder for efficient, zero-shot generation and rapid fine-tuning for customized editing.
Recently, Dragon \cite{Mou2023} and Diffeditor \cite{Mou2024} designed the energy-guided sampling with pre-trained SD, which can achieve accurate personalized editing while the backbone remains training-free. However, generalized cross-category object replacement remains challenging for these methods and is worth further exploration. 

\section{Preliminary}
\label{Preliminary}
\subsubsection{Score Distillation Sampling (SDS).}

Given input image \( \mathbf{x}_0 \), text embedding \( y \), noise \( \epsilon \sim \mathcal{N}(0, \mathbf{I}) \), timestep \( t \sim \mathcal{U}(0, 1) \), and noise estimater \( \epsilon_\phi \) parameterized by \( \phi \), the diffusion loss is defined by:
\[
\mathcal{L}_{\mathrm{diff}} = w(t) \left\| \epsilon_{\boldsymbol{\phi}}(\mathbf{x}_t, y, t) - \boldsymbol{\epsilon} \right\|_2^2,
\]
where \( w(t) \) is a weighting function, \(\mathbf{x}_t = \sqrt{\alpha_t}\mathbf{x}_0 + \sqrt{1 - \alpha_t}\epsilon\).
Classifier-free Guidance (CFG) \cite{Ho2022} achieves high-quality text-conditioned generation via guidance scale \( \omega \):
\(
\epsilon_{\boldsymbol{\phi}}^{\omega}(\mathbf{x}_t, y, t) = \omega \epsilon_{\boldsymbol{\phi}}(\mathbf{x}_t, y, t) + (1 - \omega) \epsilon_{\boldsymbol{\phi}}(\mathbf{x}_t, t).
\)
SDS \cite{Poole2022} and DDS \cite{Hertz2023} utilize the diffusion loss to optimize the image or model parameters $\theta$ towards the text condition $y$:
\[
\begin{split}
\nabla_{\theta}\mathcal{L}_{\mathrm{SDS}} &= (\epsilon_{\boldsymbol{\phi}}^{\omega}(\mathbf{x}_t, y, t) - \boldsymbol{\epsilon}) \frac{\partial \mathbf{x}_t}{\partial \boldsymbol{\theta}}, 
\\
\nabla_{\theta}\mathcal{L}_{\mathrm{DDS}} &= (\epsilon_{\boldsymbol{\phi}}^{\omega}(\mathbf{x}_t, y, t) - \epsilon_{\boldsymbol{\phi}}^{\omega}(\mathbf{\hat{x}}_t, \hat{y}, t)) \frac{\partial \mathbf{x}_t}{\partial \boldsymbol{\theta}}.
\end{split}
\]

\subsubsection{Energy-based Guidance.}
Song et al. \cite{Song2020} introduces conditional generation by decomposing score function \(\nabla_{\mathbf{x}_t}\log p(\mathbf{x}_t|\mathbf{c})\) into:
\[
\nabla_{\mathbf{x}_t}\log p(\mathbf{x}_t|\mathbf{c}) = \nabla_{\mathbf{x}_t}\log p(\mathbf{x}_t) + \nabla_{\mathbf{x}_t}\log p(\mathbf{c}|\mathbf{x}_t),
\]
The challenge lies in modeling the correction gradient \(\nabla_{\mathbf{x}_t}\log p(\mathbf{c}|\mathbf{x}_t)\).
A solution involves using energy function:
\[
p(\mathbf{c}|\mathbf{x}_t) = \frac{\exp\{-\lambda\mathcal{E}(\mathbf{c}, \mathbf{x}_t)\}}{Z},
\]
where \(\lambda\) is a positive temperature coefficient, \(Z\) is a normalizing factor, and \(\mathcal{E}(\mathbf{c}, \mathbf{x}_t)\) is an energy function measuring compatibility between condition \(\mathbf{c}\) and noised image \(\mathbf{x}_t\). Lower energy values indicate higher compatibility.
This formulation approximates the correction gradient as
\[
\nabla _{\mathbf {x}_t}\log p(\mathbf {c}|\mathbf {x}_t) \propto -\nabla _{\mathbf {x}_t}\mathcal {E}(\mathbf {c}, \mathbf {x}_t).
\]
The SDS gradient can be interpreted within this energy-based framework.
It can be decomposed into two components: \(\epsilon_{\boldsymbol{\phi}}(\mathbf{x}_t, t)\) corresponding to \(\nabla_{\mathbf{x}_t}\log p(\mathbf{x}_t)\), and \(\omega(\epsilon_{\boldsymbol{\phi}}(\mathbf{x}_t, y, t) - \epsilon_{\boldsymbol{\phi}}(\mathbf{x}_t, t))\) associated with \(\nabla_{\mathbf{x}_t}\log p(\mathbf{c}|\mathbf{x}_t)\).

\section{Method}
\label{Method}

\subsection{Energy-Guided Optimization for Image Editing}
Classifier-based methods \cite{Dhariwal2021, Zhao2022, Liu2023} use time-dependent distance measuring functions to approximate energy functions:
\[
\mathcal{E}(c, \mathbf{x}_t) \approx \mathcal{D}_\phi(c, \mathbf{x}_t, t),
\]
where \(\phi\) denotes the pre-trained parameters. \(\mathcal{D}_\phi(c, \mathbf{x}_t, t)\) denotes the distance between condition \(c\) and noised latent \(\mathbf{x}_t\).
While off-the-shelf pre-trained networks (e.g., classification, segmentation) can be used, they often rely on one-step denoising approximations \cite{Yu2023}, leading to inaccuracies, especially in the early stage of generation.
We propose using the pre-trained diffusion model itself as energy function to leverage its comprehensive image prior knowledge:
\[
\mathcal{E}(\mathbf{c}, \mathbf{\tilde{x}}_0)) \approx \mathbb{E}_{p(\mathbf{\tilde{x}}_t)|\mathbf{\tilde{x}}_0))}[\mathcal{D}_{\theta}(\mathbf{c}, \mathbf{\tilde{x}}_t)],
\]
measuring the distance between the target image \(\tilde{\mathbf{x}}_0\) and the condition \(\mathbf{c}\) at various noise levels \(t\), based on the intuition that the distance between a noised image \(\tilde{\mathbf{x}_t}\) and the condition \(\mathbf{c}\) reflects the distance between its corresponding target image \(\tilde{\mathbf{x}}_0\) and the same condition \(\mathbf{c}\).

Our training-free approach (see Fig. \ref{pipeline}) regards image editing as a conditional optimization problem.
Instead of relying on iterative sampling, we directly update the target image \(\tilde{\mathbf{x}}_0\) under the guidance from both text and image energy,
which can be formulated as:
\[
\begin{split}
\mathcal{E}(\mathbf{c}_{txt}, \mathbf{c}_{img}, \tilde{\mathbf{x}}_0) &\approx \eta_{1}\mathbb{E}_{p(\mathbf{\tilde{x}}_t|\tilde{\mathbf{x}}_0)}[\mathcal{D}_{\theta}^t(\mathbf{c}_{txt}, \mathbf{\tilde{x}}_t)] \\ 
&+ \eta_{2}\mathbb{E}_{p(\mathbf{\tilde{x}}_t|\tilde{\mathbf{x}}_0)}[\mathcal{D}_{\theta}^i(\mathbf{c}_{img}, \mathbf{\tilde{x}}_t))],
\end{split}
\]
where \(\eta_1\) and \(\eta_2\) are the weight coefficients. 
The text energy provides high-level conceptual guidance particularly in the early stage of optimization, ensuring that the editing result aligns with the text prompt.
While the image energy provides fine-grained control in the later stage by enhancing appearance details.

\subsubsection{Text Energy.}
Preliminarily, distillation loss proposed by SDS and DDS can be regarded as text energy.
We further extend the formulation to incorporate with negative prompts:
\[
\begin{split}
\mathcal{D}_{\theta}^t(\mathbf{c}_{txt},\mathbf{\tilde{x}}_t)) =  ||&\epsilon_{\boldsymbol{\phi}}^{\omega}(\mathbf{\tilde{x}}_t,y_{ref},t)-\epsilon_{\boldsymbol{\phi}}^{\omega}(\mathbf{x}_t^{src},y_{src},t)||_2^2+\\ ||&\epsilon_{\boldsymbol{\phi}}(\mathbf{\tilde{x}}_t,t)-\epsilon_{\boldsymbol{\phi}}(\mathbf{\tilde{x}}_t,y_{neg},t) ||_2^2.
\end{split}
\]
where \(y_{src}\), \(y_{ref}\) and \(y_{neg}\) represents two positive prompt for source and target objects respectively, along with an extra negative prompt.
By leveraging these prompts, we achieve precise control over object attributes.
However, it brings a trade-off between quality and fidelity, as negative prompts may cause deviations from the target prompt, leading to visible artifacts.
Thus we assign a smaller weight to the energy gradient from negative prompt.

Given that text prompts are usually ambiguous and lack the detailed description necessary for precise guidance in personalized image editing, they may conflict with the details recovered through image energy.
The IP-Adapter \cite{Ye2023} is compatible with the text energy function and can further enhance the gradient precision for editing with image tokens as prompts.
However, the image background can be inadvertently altered after introducing the IP-Adpater since background information is also captured by image tokens. We employ Gradient Scaling (GS) to adjust the text energy gradient. Specifically, for the text energy function, we use Background Gradient Truncation to limit the energy gradient outside the mask \(M_{ori}\): 
\(
\nabla_{x_0}\mathcal{D}_\theta^t \odot M_{ori}.
\)

\begin{figure}[t]
  \centering
  \includegraphics[width=\linewidth]{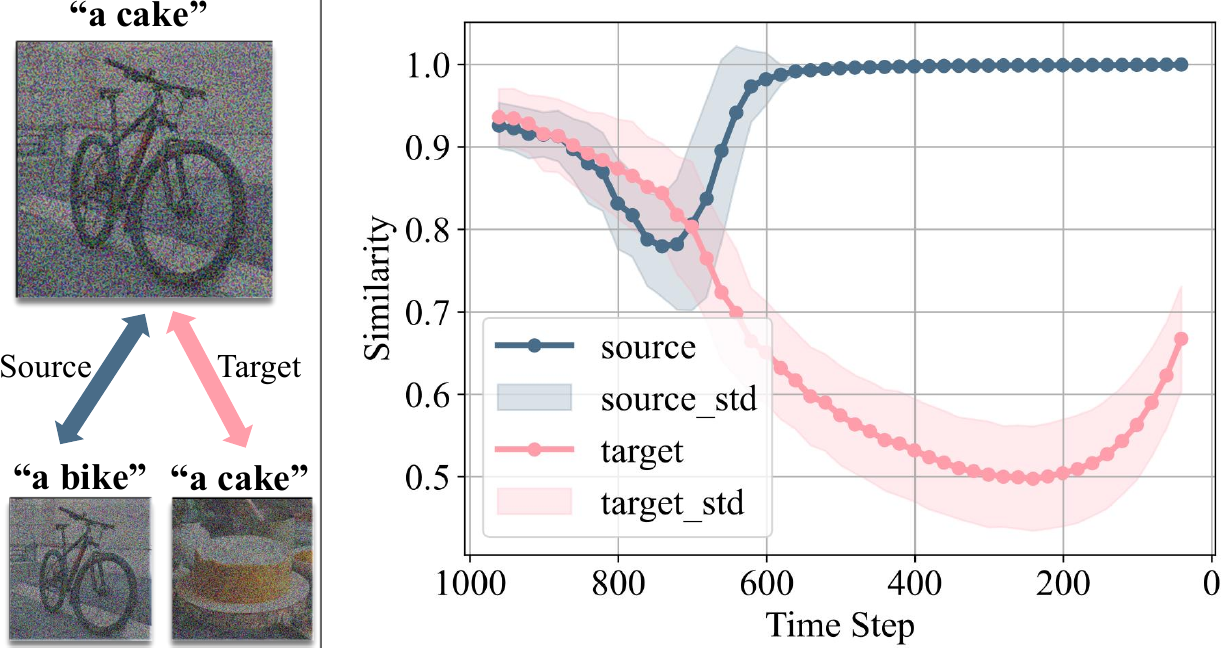}
  \caption{Visualization of the feature similarity. Given two text-image pairs, referred as source and target. We query the source image with the target text and compute the feature similarity with the source and target under noise addition to different times \(t\). We analyze 140 images from different categories and calculate the mean and standard deviation of the feature similarity.}
  \label{observation}
\end{figure}

\subsubsection{Image Energy.}

Image energy focuses primarily on the transfer of reference features.
At each timestep \(t\), the UNet denoiser \(\epsilon_\phi\) is utilized to extract intermediate features \(\mathcal{F}^{opt}_t\) from the edited image \(\mathbf{\tilde{x}}_t\), so do the reference features \(\mathcal{F}^{ref}_t\) from the reference image \(\mathbf{x}_t^{ref}\).

Following the methodology outlined in DIFT \cite{Tang2023}, \(\mathcal{F}^{opt}_t\) and \(\mathcal{F}^{ref}_t\) preserve the high-level semantic consistency required for precise point-to-point correspondence.
To ensure coherent edits across the image, we employ two binary masks to constrain the region for editing.
Within the mask, we focus on transferring the features by identifying the points in \(\mathcal{F}^{ref}_t\) that have the shortest distance to the corresponding points in 
\(\mathcal{F}^{opt}_t\). 

Calculating the shortest distance ensures that the transferred features from the reference image maintain their semantic alignment in the optimized image, resulting in precise and contextually consistent edits.
The distance metric \(d(\cdot, \cdot)\) can be but not limited to Euclidean distance:
\begin{equation}
p_{o}=\mathop{\rm argmin}_{p \in M_{ori}}{d}(\mathcal{F}^{ref}_t[p_{r}],\mathcal{F}^{opt}_t[p]).
\label{eq:match}
\end{equation}

Here, \(\mathcal{F}^{opt}_t[p]\) and \(\mathcal{F}^{ref}_t[p_r]\) represent the feature vectors at the point \(p\in \mathcal{M}_{ori}\) and \(p_r\in \mathcal{M}_{ref}\).
After finding the matching point \(p_o\) for each \(p_r\). The image energy function is thus defined to minimize the point-to-point feature distance inside the mask:
\[
\mathcal{D}_{\theta}^i(\mathbf{c}_{img},\tilde{\mathbf{x}}_t)= \sum_{p_r \in M_{ref}}||\mathcal{F}^{opt}[p_o] - \mathcal{F}^{ref}[p_r]||^2_2.
\]

Unlike text energy which provides the coarse semantic guidance, we aim for image energy to provide fine-grained transition control upon the masked regions.
If the gradient update is restricted within the region, it may reduce editing flexibility and result in visible artifacts along the object boundary.
Therefore, another Gradient Scaling operation is introduced, we apply Background Gradient Smoothing to ensure a more natural transition at the boundary: 
\(\nabla_{x_0}\mathcal{D}_\theta^i \odot M_{ori} + (\nabla_{x_0}\mathcal{D}_\theta^i * k_{s}) \odot (1 - M_{ori}) \), where $k_{s}$ is the Gaussian smoothing kernel.

\subsection{Latent Space Content Composition}

Although conditional energy function optimization can effectively integrate the target object into the source image, there still exists inconsistency in appearance due to the information loss and large domain gaps  
in certain scenarios.
It demands a more efficient feature transferring method to achieve higher consistency.

Notice that the feature matching results in Eq.~\ref{eq:match} and the structural information become stable after some iterations, we design a Content Composition operation accordingly.  
This operation directly manipulates latent variables in a copy-paste manner, transferring features between the matching point pairs from the reference image to \(\tilde{x}_0\) within the mask.
Given a point pair $(p_o, p_r)$ in Eq.~\ref{eq:match}, we have:
\[
\tilde{x}_0[p_{o}] \leftarrow x^{ref}_0[p_{r}].
\]

By directly transferring in the latent space, local appearance information from the target is transferred to the edited results. Moreover, unlike the one-step copy-paste in the pixel space, we apply the operation at an interval during optimization. This allows the subsequent optimization with energy guidance to repair the integrity and semantics of the result.

\begin{figure*}[t]
\centering
    \includegraphics[width=0.86\textwidth]{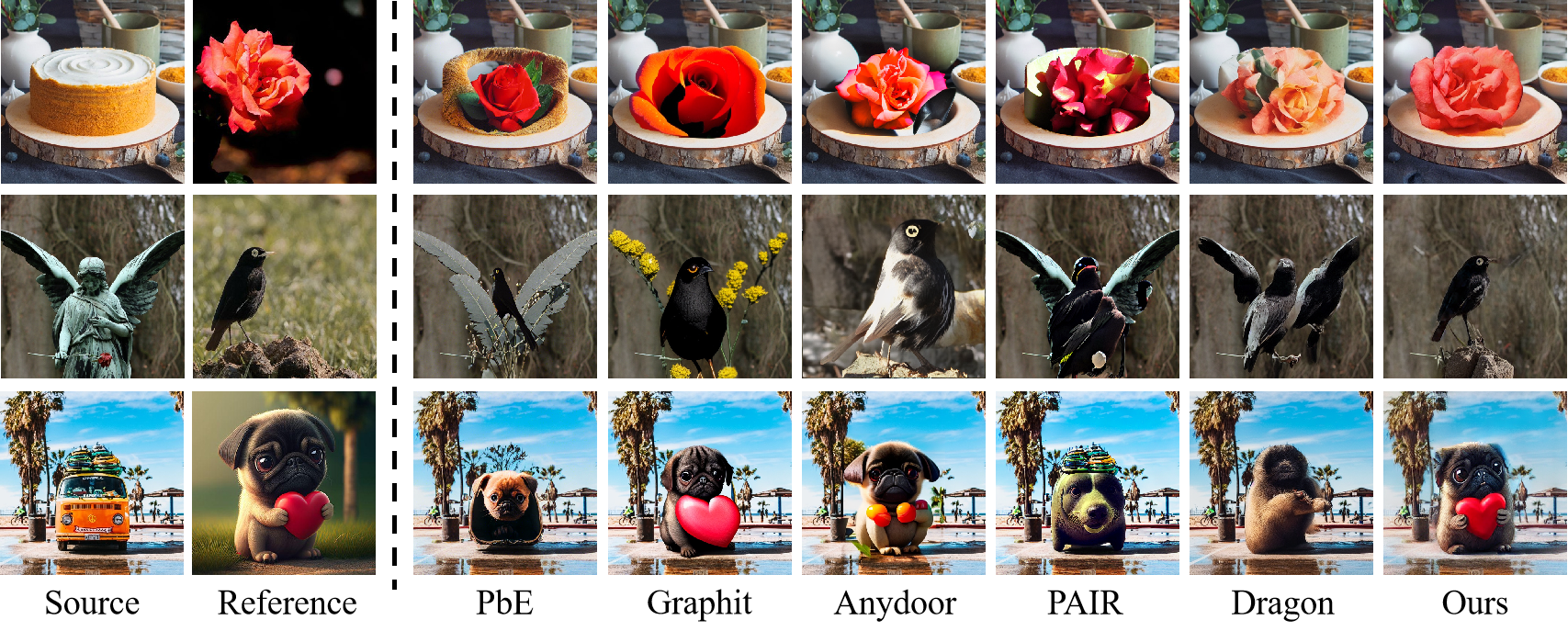}
    \caption{Qualitive results of cross-class replacement. The source object and the target object are sampled from different classes.}
    \label{across categories}
\end{figure*}

\subsection{Coarse-to-fine Optimization}

In the original SDS and DDS methods, optimization is performed by randomly sampling timesteps per iteration.
However, this strategy can lead to blurry edited results when optimizing for the energy function $\mathcal{E}$ due to instability in feature matching~\cite{Tang2023}.
Furthermore, the high similarity between the edited and source images in early optimization iterations makes the feature matching inaccurate, causing misleading image energy guidance \(D_{\theta}^i\).

To address these limitations, we introduce a coarse-to-fine scheduling strategy, which narrows the timestep \(t\) to a pre-defined sequence \( \{t_1, t_2, t_3, \dots, t_N\}\), satisfying
\(
t_1 > t_2 > t_3 > \dots > t_N
\). 
Larger \(t\) captures coarse-grained concepts, while smaller \(t\) focuses on fine-grained details, resulting in more effective edits.

We emphasize text energy during early optimization stages with large noise for narrowing the semantic gaps.
As optimization progresses, image energy becomes increasingly relevant to the editing result.
Fig. \ref{observation} illustrates the variation in feature similarity between edited and source images with respect to timestep \(t\).
At large timestep \(t>700\), the edited images resemble target images, indicating the dominant influence of text energy.
After timestep \(t=600\), the similarity between edited and source images surpasses that with the target image, indicating a shift towards image feature dominance.
Therefore, we opt to image energy gradients to guide detail restoration during the middle and later stages of optimization.

This coarse-to-fine strategy effectively employs text and image energy functions: large time steps (early stage) for establishing structure via text energy, medium time steps (middle stage) for refining features with increasing image energy influence, and small time steps (later stage) for fine details jointly guided by text-image energy. 

\section{Experiments}
\label{Experiments}
\subsection{Settings}
\subsubsection{Benchmarks.}
We evaluated our method using established benchmarks. For object swapping tasks, we utilized DreamEditBench \cite{Li2023b}, which features 22 themes aligned with the DreamBooth framework \cite{Ruiz2023}. We performed two-by-two exchanges using 10 images from the same theme but different environments within DreamEditBench. Additionally, we selected 50 images from PIE-Bench \cite{Ju2024}, representing distinct conceptual categories, and paired them randomly for object swapping.

\subsubsection{Implementation Details.}
We implemented our proposed method using the Stable Diffusion 1.5 model as the pre-trained text-to-image diffusion model. All experiments were conducted on images with a resolution of 512 x 512 pixels, striking a balance between image quality and computational efficiency. The number of optimization steps is set to 50 for all experiments. More details can be found in Appendix A.

\begin{table}[t]
\centering
\begin{tabular}{@{}lcccc@{}}
\toprule
Method  & DINO ($\uparrow$) & CLIP-T  ($\uparrow$) & CLIP-I ($\uparrow$) \\ \midrule
PbE  & 47.666 & 27.149 & 71.322 \\
Anydoor  & 57.093 & 26.410 & 72.133  \\ 
PAIR   & 48.786 & 25.901 & 68.575  \\
Dragon  & 48.176 & 26.361 & 69.949 \\
EGO-Edit(Ours)    & \textbf{62.749} & \textbf{28.764} & \textbf{76.624}
\\
\bottomrule
\end{tabular}
\caption{Quantitative comparison results. DINO and CLIP scores are reported to evaluate the quality of replacement.}
\label{tab:object_swap}
\end{table}

\subsubsection{Evaluation Metrics.}
To comprehensively assess the performance of our proposed method, we employed a diverse set of evaluation metrics: CLIP Score (text and image) \cite{Radford2021}  and DINO Score \cite{Caron2021}. 
These metrics effectively reflect the similarity between the generated region and the target object or the matching degree with the text. The user study is presented in Appendix B.
\begin{figure}[t]
\centering
    \includegraphics[width=\linewidth]{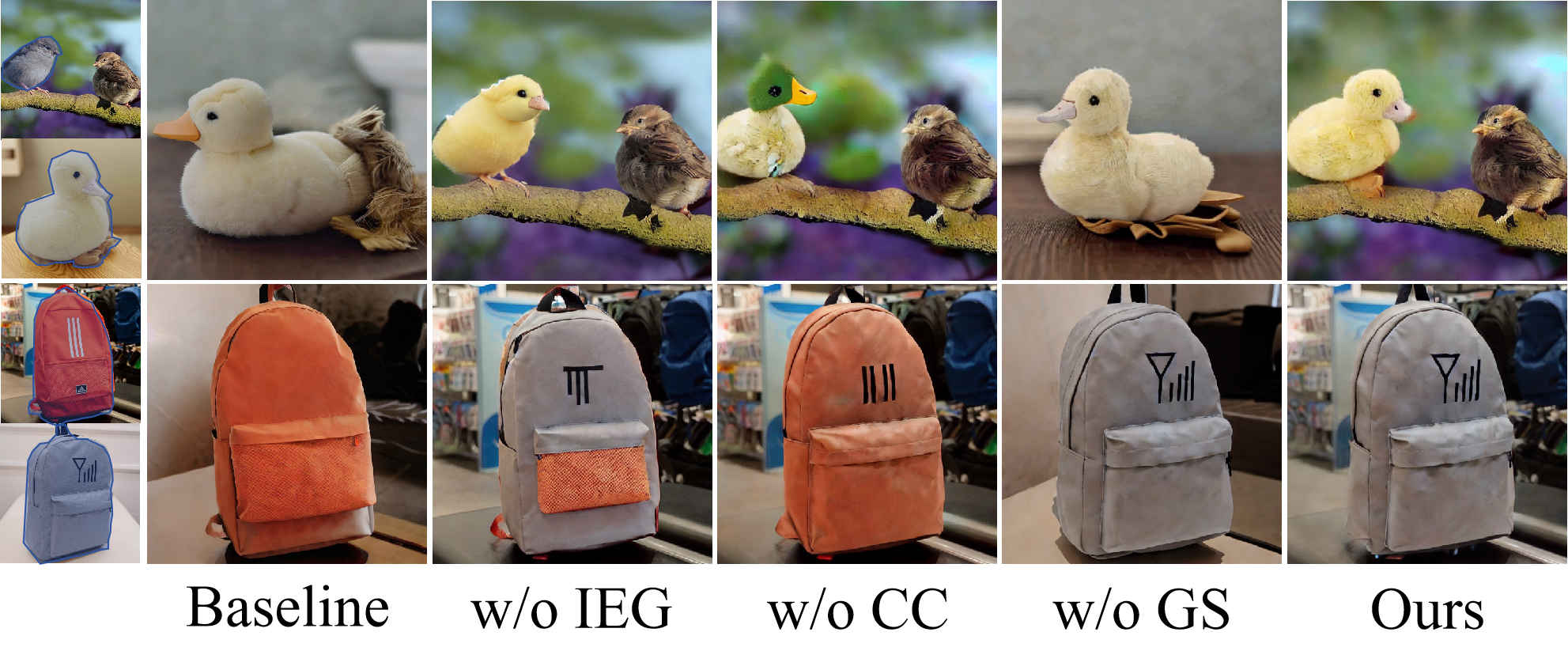}
    \caption{Ablation on the components. The baseline uses the text guidance and the IP-adapter. ``IEG": image-energy guidance. ``CC": content composition. ``GS": gradient scaling.}
    \label{components_ablation}
\end{figure}

\begin{figure*}[t]
\centering
    \includegraphics[width=0.86\textwidth]{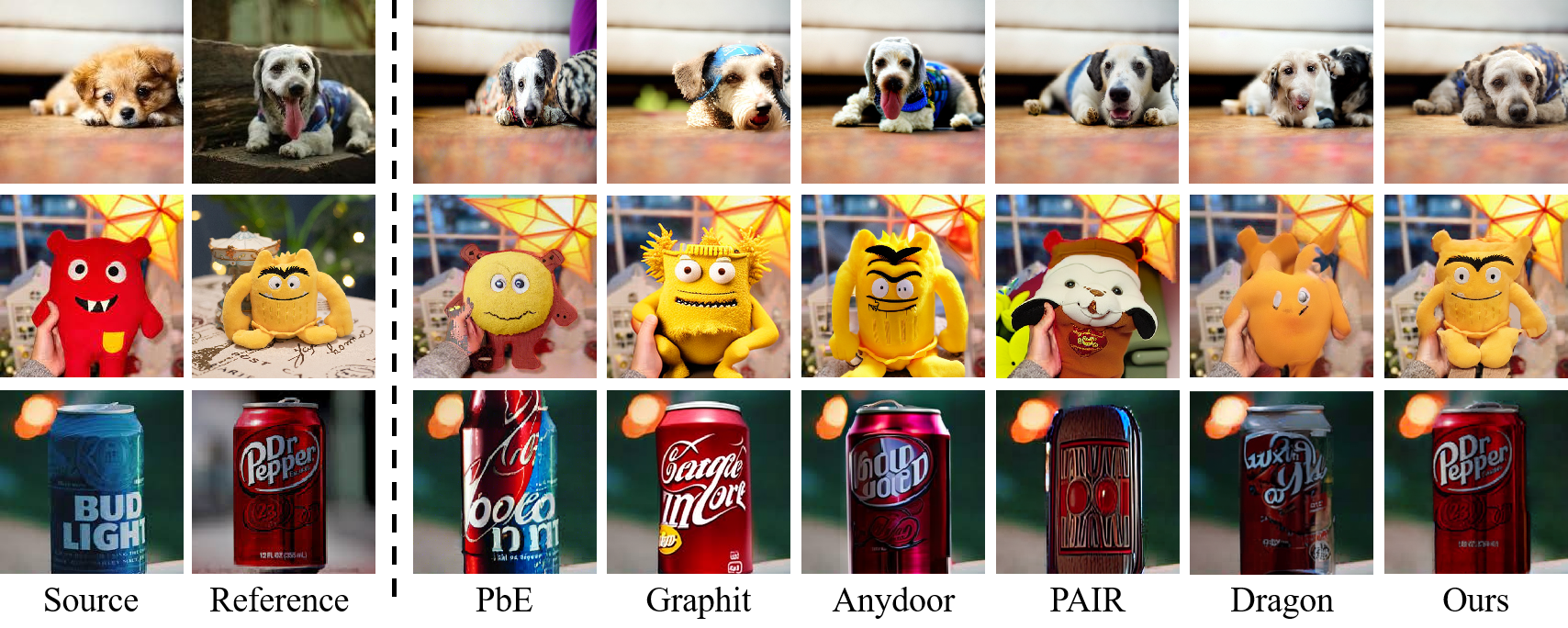}
    \caption{Qualitative results of in-class replacement. The target object in the reference image belongs to the source object class.}
    \label{same categories}
\end{figure*}

\subsection{Comparison with Previous Methods}
We present the quantitative results of our method and the competing baselines on the customized image swapping tasks in Table \ref{tab:object_swap}. Our approach demonstrates superior performance in both DINO and CLIP metrics, indicating higher fidelity in the edited images compared to existing methods. 

We provide a qualitative comparison of cross-class replacement between the proposed method and competing approaches in Fig. \ref{across categories}. When there is a significant size difference between the source object and the target object (see second row), inpainting-based methods (PbE \cite{Yang2023a}, Graphit \cite{Graphit}, Anydoor \cite{Chen2024a}) often introduce irrelevant details within the mask area. Appearance editing-based methods (PAIR \cite{Goel2023}, Dragon \cite{Mou2023}) tend to fill the entire mask area with the reference object. In contrast, our method follows the size of the reference image and can make a smooth transition from the source image.
In Fig. \ref{same categories}, we further visualize the results of in-class replacement. Our method can transfer the reference image features while maintaining the source image pose.
More applications are shown in Appendix C.

\subsection{Ablation Study}
\subsubsection{Components.}
To understand the contributions of different components in our method, we conducted an ablation study, with the results presented in Table \ref{tab:core_components_quantitative_ablation}.
For a clearer comparison, Fig. \ref{components_ablation} illustrates the effects of each component. The last column presents results from our method, while the preceding three columns demonstrate the impact of systematically removing key components.
Without IEG, the generated images lost distinctive identity features, maintaining only coarse semantic consistency.
Excluding Content Composition (CC) led to significant issues: while image energy guidance transferred relevant features, it often missed key details and inadequately preserved color information, causing noticeable degradation of fine details like the duck's head color (row 1) and the bag's logo (row 2).
Removing Gradient Scaling (GS) results in the influence of semantic guidance on the entire image, which adversely impacts editing quality.

\begin{table}[t]
\centering
\begin{tabular}{lcccc}
\toprule
Method  & DINO ($\uparrow$) & CLIP-T  ($\uparrow$) & CLIP-I ($\uparrow$) \\ \midrule
w/o IEG    & 56.510 & 28.660 & 75.664 \\
w/o CC & 57.531 & 28.753 & 75.457  \\
EGO-Edit & \textbf{62.749} & \textbf{28.764} & \textbf{76.624} \\
\bottomrule
\end{tabular}
\caption{Quantitative ablation results of components. We compare the results without image-energy guidance (IEG) or content composition (CC).}
\label{tab:core_components_quantitative_ablation}
\end{table}

\subsubsection{Timestep Selection for Content Composition.}

\begin{figure}[t]
\centering
    \includegraphics[width=0.9\linewidth]{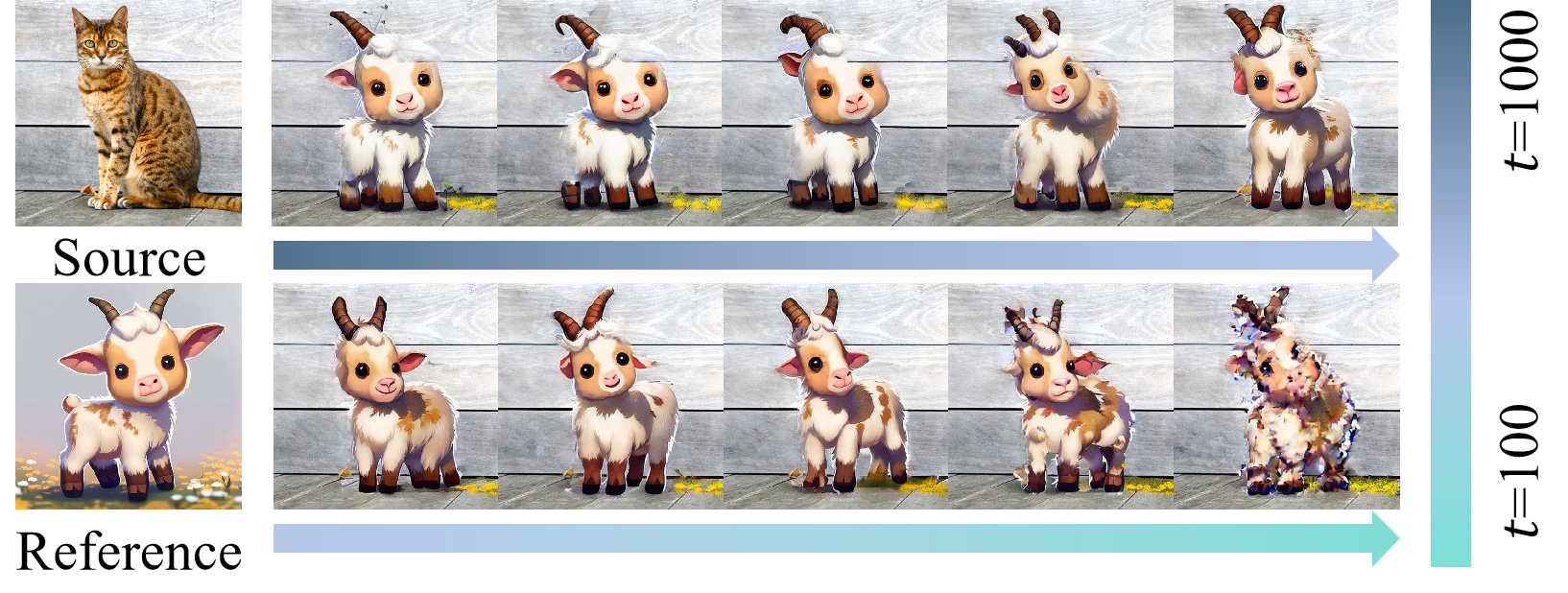}
    \caption{Ablation on the insertion condition $t$ for CC. We start CC when the timestep is $t$ and stop it with the timestep $t-100$ in optimization.}
    \label{ablation_on_t}
\end{figure}

We investigated the key factors for the insertion of the content composition. In Fig. \ref{ablation_on_t}, we compare edited results with different timesteps \(t\) to insert CC. We concluded that performing content composition early in the optimization process (upper left) results in generated images that exhibit high consistency with the reference image in terms of pose and relative position. However, early content composition can lead to excessive adherence to the reference, potentially sacrificing flexibility in the final output.
Conversely, conducting content composition at the late stage of optimization (lower right) fails to adequately correct artifacts caused by discrete matching, leading to image blurriness and a lack of coherent details. It appears that late iterations (smaller $t$) do not provide sufficient opportunity to refine the transferred features, resulting in lower quality in the generated image.
Optimal results are achieved when the content composition is applied at the medium stages of the optimization process. This choice balances the trade-off between adherence to the reference image and flexibility in the final output.

\subsubsection{Optimization Strategy.}

\begin{figure}[t]
\centering
    \includegraphics[width=\linewidth]{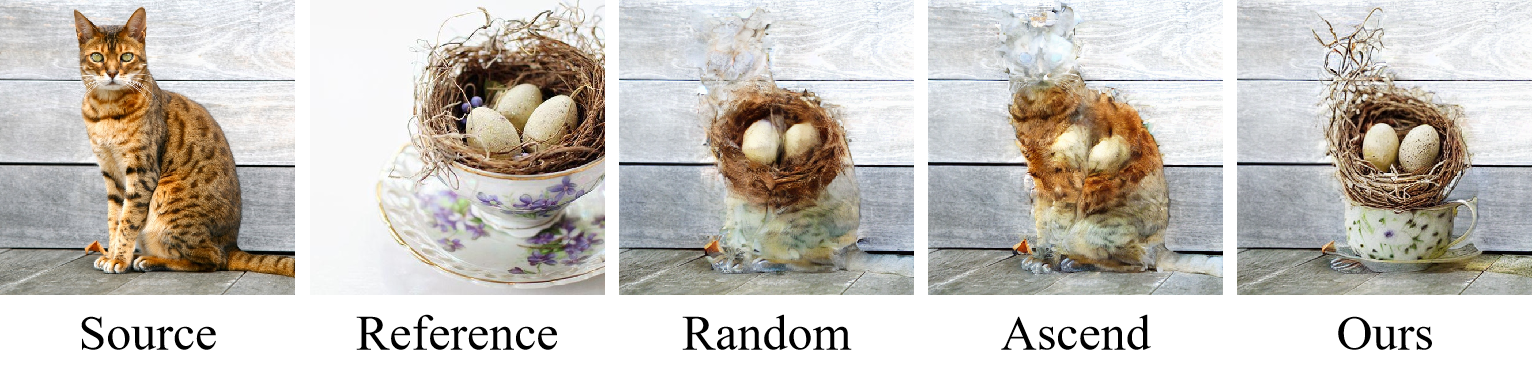}
    \caption{Ablation on the optimization schedule of timesteps. We compare three schedules of time steps during optimization: random, ascending order, and descending order. The descending schedule in our method performs best on the consistency of both details and semantics.}
    \label{optimizer_strategy}
\end{figure}

We compare three optimization schedules of timesteps, namely random, ascending, and descending in Fig. \ref{optimizer_strategy}. 
The visual comparison reveals that the random schedule suffer from noticeable blurriness. This is due to the inherent instability of the random strategy in feature matching, resulting in inconsistent feature point alignment across optimization iterations. 
The ascending schedule optimizes the fine-grained appearance first (start from small timestep $t$). However, the semantic gap at the early stage causes a shape mismatch with the target object. The mismatch makes the edited result exhibit obvious shape characteristics of the source object (cat).
We employ the descending schedule in the coarse-to-fine optimization that aligns the coarse shape information with the target first. This allows more accurate and stable matching for refining the appearance details and achieves better quality.

\section{Conclusion}
\label{Conclusion}
This paper introduces a paradigm shift in personalized image editing through the innovative application of pretrained diffusion models. We address the limitations of existing methods by proposing a training-free and inversion-free approach that harnesses the conditional optimization of latent codes, guided by reference images and text.
Our method minimizes feature discrepancies between edited and reference images, ensuring a seamless integration of personalized content. Extensive experiments have shown that our method has achieved good results in both in-class and cross-class customized object replacement.

\section{Acknowledgements}
\label{Acknowledgements}
This work is supported in part by National Science Foundation for Distinguished Young Scholars under Grant 62225605, Zhejiang Provincial Natural Science Foundation of China under Grant LD24F020016, Natural Science Foundation of Shanghai under Grant 24ZR1425600, Project 12326608 supported by NSFC, "Pioneer" and "Leading Goose" R\&D Program of Zhejiang (No. 2024C01020), National Natural Science Foundation of China under Grant No.62441602, the Ningbo Science and Technology Innovation Project (No.2024Z294), and the Fundamental Research Funds for the Central Universities.

\bibliography{aaai25}

\end{document}